\documentclass{article}

\usepackage{color}
\usepackage{amsmath}
\usepackage{amssymb}
\usepackage{multirow, varwidth}
\usepackage{epsfig}
\usepackage{verbatim}
\makeatletter
\newif\if@restonecol
\makeatother
\usepackage{xr}
\usepackage[amsmath,thmmarks]{ntheorem}
\usepackage{flushend}

\theorembodyfont{\normalfont}
\theoremseparator{.}

\theoremstyle{nonumberplain}

\DeclareRobustCommand\onedot{\futurelet\@let@token\@onedot}
\def\onedot{.}

\usepackage{microtype}
\usepackage{graphicx}
\usepackage{subfigure}
\usepackage{booktabs}

\usepackage{epsfig}
\usepackage{amsmath}
\usepackage{amssymb}
\usepackage{paralist}
\usepackage{algorithm}
\usepackage{algorithmic}
\usepackage{nicefrac}

\usepackage{pifont}%
\newcommand{\cmark}{\ding{51}}
\newcommand{\xmark}{\ding{55}}

\usepackage[dvipsnames]{xcolor}

\usepackage{hyperref}

\usepackage[accepted, nohyperref]{icml2018}

\icmltitlerunning{Video Prediction with Appearance and Motion Conditions}

\begin{document}

\twocolumn[
\icmltitle{Video Prediction with Appearance and Motion Conditions}

\icmlsetsymbol{equal}{*}

\begin{icmlauthorlist}
\icmlauthor{Yunseok Jang}{umich,snu}
\icmlauthor{Gunhee Kim}{snu}
\icmlauthor{Yale Song}{msair}
\end{icmlauthorlist}

\icmlaffiliation{umich}{University of Michigan, Ann Arbor}
\icmlaffiliation{snu}{Seoul National University}
\icmlaffiliation{msair}{Microsoft AI \& Research}

\icmlcorrespondingauthor{Yunseok Jang}{yunseokj@umich.edu}
\icmlcorrespondingauthor{Gunhee Kim}{gunhee@snu.ac.kr}
\icmlcorrespondingauthor{Yale Song}{yalesong@microsoft.com}

\icmlkeywords{video understanding, generative adversarial network, computer vision, triplet loss}

\vskip 0.3in
]

\printAffiliationsAndNotice{This work was done at Yahoo Research during summer internship.}  %

\begin{abstract}
Video prediction aims to generate realistic future frames by learning dynamic visual patterns. One fundamental challenge is to deal with future uncertainty: How should a model behave when there are multiple correct, equally probable future? We propose an Appearance-Motion Conditional GAN to address this challenge. We provide appearance and motion information as conditions that specify how the future may look like, reducing the level of uncertainty. Our model consists of a generator, two discriminators taking charge of appearance and motion pathways, and a perceptual ranking module that encourages videos of similar conditions to look similar. To train our model, we develop a novel conditioning scheme that consists of different combinations of appearance and motion conditions. We evaluate our model using facial expression and human action datasets and report favorable results compared to existing methods.
\end{abstract}

\section{Introduction}
\label{sec:introduction}

Video prediction is concerned with generating high-fidelity future frames given past observations by learning dynamic visual patterns from videos. It is a promising direction for video representation learning because the model will have to learn to disentangle factors of variation based on complex visual patterns, i.e., how objects move and deform over time, how scenes change as the camera moves, how background changes as the foreground objects move, etc. While the recent advances in deep generative models~\cite{kingma-arxiv13,goodfellow-nips14} have brought a rapid progress to image generation~\cite{radford-iclr16,isola-cvpr17,zhu-iccv17}, relatively little progress has been made in video prediction. We believe this is due in part to future uncertainty~\cite{walker-eccv16}, making the problem somewhat ill-posed and evaluation difficult.

Previous work has addressed the uncertainty issue in several directions. One popular approach is \textit{learning to extrapolate} multiple past frames into the future~\cite{srivastava-icml15, mathieu-iclr16}. This helps reduce uncertainty because input frames act as conditions that constrain the range of options for the future. However, when input frames are not sufficient statistics of the future, which is often the case with just a few frames (e.g., four in \cite{mathieu-iclr16}), these methods suffer from blurry output caused by future uncertainty. Recent methods thus leverage \textit{auxiliary information}, e.g., motion category labels and human pose, along with multiple input frames~\cite{finn-nips16, villegas-icml17, walker-iccv17}. Unfortunately, these methods still suffer from motionless and/or blurry output caused by the lack of clear supervision signals or suboptimal solutions found by training algorithms.

In this work, we propose an Appearance-Motion Conditional Generative Adversarial Network (AMC-GAN). Unlike most existing methods that learn from \textit{multiple} input frames~\cite{srivastava-icml15, mathieu-iclr16,finn-nips16,villegas-icml17,liang-iccv17}, which contain both appearance and motion information, we instead disentangle appearance from motion, and learn from a \textit{single} input frame (appearance) and auxiliary input (motion). This allows our model to learn different factors of variation more precisely. Encoding motion with an auxiliary variable allows our model to \textit{manipulate} how the future would look like; with a simple change of the auxiliary variable, we can make a neutral face happy or frown, or make a neutral body pose perform different gestures.  

Training GANs is notoriously difficult~\cite{salimans-nips16}. We develop a novel conditioning scheme that constructs multiple different combinations of appearance and motion conditions -- including even the ones that are \textit{not} part of the training samples -- and specify constraints to the learning objective such that videos generated under different conditions all look plausible. This makes the model generate videos under conditions \textit{beyond what is available in the training data} and thus work much harder to satisfy the constraints during training, improving the generalization ability. In addition, we incorporate perceptual triplet ranking into the learning objective so that videos with similar conditions look more similar to each other than the ones with different conditions. This mixed-objective learning strategy helps our model find the optimal solution effectively.

One useful byproduct of our conditional video prediction setting is that we can design an \textit{objective} evaluation methodology that checks whether generated videos contain the likely content as specified in the input condition. This is in contrast to the traditional video prediction setting where there is no expected output, other than it being plausibly looking~\cite{vondrick-nips16}. We design an evaluation technique where we train a video classifier on real data with motion category labels and test it on generated videos. We also perform qualitative analysis to assess the visual quality of the output, and report favorable results on the MUG facial expression dataset~\cite{aifanti-wiamis10} and the NATOPS human action dataset~\cite{song-fg11}.

To summarize, our contributions include:
\begin{compactitem}
\item We propose AMC-GAN that can generate multiple different videos from a single image by manipulating input conditions. The code is available at \href{http://vision.snu.ac.kr/projects/amc-gan}{\color{magenta}http://vision.snu.ac.kr/projects/amc-gan}.
\item We develop a novel conditioning scheme that helps the training by varying appearance and motion conditions. 
\item We use perceptual triplet ranking to encourage videos of similar conditions to look similar. To our best knowledge, this has not been explored in video prediction.
\end{compactitem}

\section{Related Work}
\label{sec:related_works}

\textbf{Future Prediction:} Early work proposed to use the past observation to predict certain representation of the future, e.g., object trajectory~\cite{walker-cvpr14}, optical flow~\cite{walker-iccv15}, dense trajectory features~\cite{walker-eccv16}, visual representation~\cite{vondrick-cvpr16}, and human poses~\cite{chao-cvpr17}. Our work is distinct from this line of research as we aim to predict future frames rather than certain representation of the future.

\textbf{Video Prediction:} \citet{ranzato-arxiv14} proposed a recurrent neural network that predicts a target frame composed of image patches (akin to words in language). \citet{srivastava-icml15} used a sequence-to-sequence model to predict future frames. Early observations in video prediction have shown that predicted frames tend to be blurry~\cite{mathieu-iclr16,finn-nips16}. One primary reason for this is future uncertainty~\cite{walker-eccv16,xue-nips16}; there could be multiple correct, equally probable next frames given the previous frames. This observation has motivated two research directions: using adversarial training to make the predicted frames look realistic, and using auxiliary information as conditions to constrain what the future may look like. Our work is closely related to both directions as we perform conditional video prediction with adversarial training. Below we review the most representative work in the two research directions.

\textbf{Adversarial Training:} Recent methods employ adversarial training to encourage predicted frames to look realistic and less blurry. Most work differ by the design of the discriminator: \citet{villegas-icml17} use an appearance discriminator, \citet{mathieu-iclr16,villegas-iclr17,vondrick-nips16,walker-iccv17} use a motion discriminator, and \citet{liang-iccv17,tulyakov-arxiv17} use both. \citet{vondrick-nips16} use a motion discriminator based on a 3D CNN; \citet{walker-iccv17} adopt the same motion discriminator. Our motion discriminator is similar to theirs, but differ by the use of conditioning variables. \citet{liang-iccv17} define two discriminators: an appearance discriminator that inspects each frame, and a motion discriminator that inspects an optical flow image predicted from each consecutive frames. Our work also employs dual discriminators, but we do not require optical flow information. 

\textbf{Conditional Generation:} Most approaches in video prediction use multiple frames as input and predict future frames by \textit{learning to extrapolate}~\cite{ranzato-arxiv14,srivastava-icml15,mathieu-iclr16,villegas-iclr17,liang-iccv17}. We consider these methods related to ours because multiple frames essentially provide appearance and motion conditions. Some of these work, similar to ours, decompose input into appearance and motion pathways and handle them separately~\cite{villegas-iclr17,liang-iccv17}. Our work is, however, distinct from all the previous methods in that we do not ``learn to extrapolate''; rather, we learn to predict the future from \textit{a single frame} so the resulting video faithfully contains motion information provided as an auxiliary variable. This latter aspect makes our work unique because, as we show later in the paper, it allows our model to \textit{manipulate} the future depending on motion input.

For predicting future frames containing human motion, some methods estimate body pose from input frames, and decode input frames (appearance) and poses (motion) into a video~\cite{villegas-icml17,walker-iccv17}; these methods do \textit{video prediction by pose estimation}. Pose information is attractive because they are low-dimensional. Our work also uses a motion condition that is of low-dimensional, but is more flexible because we work with generic keypoint statistics (e.g., location and velocity); we show how we encode motion information in Section~\ref{sec:experiments}.

Several approaches provide auxiliary information as conditioning variables. \citet{finn-nips16} use action and state information of a robotic arm. \citet{oh-nips15} use Atari game actions. \citet{reed-icml16} propose text-to-image synthesis; \citet{marwah-iccv17} propose text-to-video prediction. These methods, similar to ours, can \textit{manipulate} how the output may look like, by changing the auxiliary information. Thus, we empirically compare our method with \citet{finn-nips16}, \citet{mathieu-iclr16} and \citet{villegas-iclr17} and report improved performance. 

Lastly, different from all above mentioned work, we incorporate a perceptual ranking loss~\cite{wang-cvpr14,gatys-nips15} to encourage videos that share the same appearance/motion conditions to look similar than videos that do not. Our work is, to the best of our knowledge, the first to use this constraint in the video prediction setting.

\section{Approach}
\label{sec:approach}

\begin{figure}
    \centering
    \vspace{-4pt}
    \includegraphics[width=.7\linewidth]{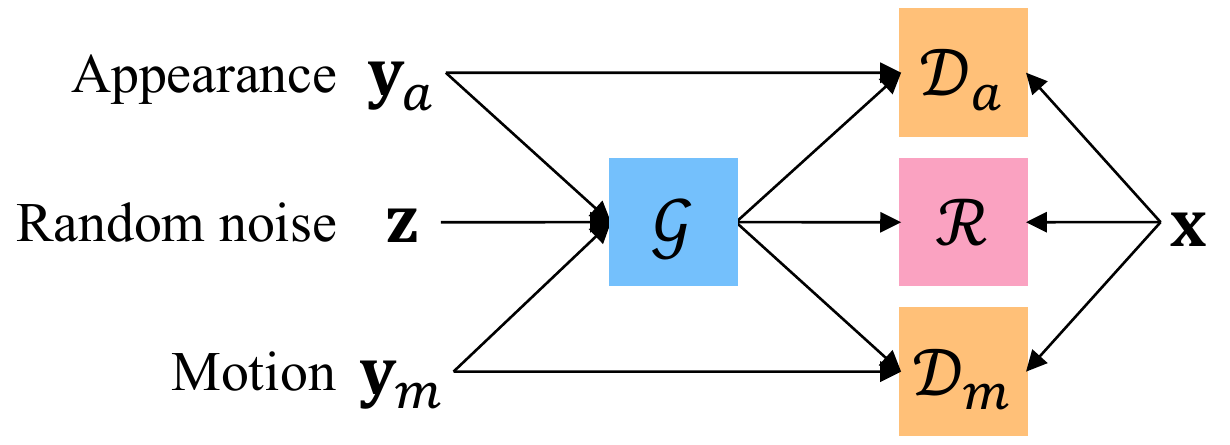}
    \vspace{-10pt}
    \caption{Our AMC-GAN consists of a generator $\mathcal{G}$, two discriminators each taking charge of appearance $\mathcal{D}_{a}$ and motion $\mathcal{D}_{m}$ pathways, and a perceptual ranking module $\mathcal{R}$.}
    \label{fig:our_model}
    \vspace{-2pt}
\end{figure}

Our goal is to generate a video given an appearance and motion information. We formulate this as learning the conditional distribution $p(\mathbf{x} | \mathbf{y})$ where $\mathbf{x}$ is a video and $\mathbf{y}=[\mathbf{y}_{a},\mathbf{y}_{m}]$ is a set of conditions known to occur. We define two conditioning variables, $\mathbf{y}_{a}$ and $\mathbf{y}_{m}$, that encode appearance and motion information, respectively. 

We propose an Appearance-Motion Conditional GAN, shown in Figure~\ref{fig:our_model}. The generator $\mathcal{G}$ seeks to produce realistic future frames. We denote a generated video by $\hat{\mathbf{x}}|_{\mathbf{y}} = \mathcal{G}(\mathbf{z}|\mathbf{y})$, where $\mathbf{z}$ is random noise. The two discriminator networks, on the other hand, attempt to distinguish the generated videos from the real ones: $\mathcal{D}_{a}$ checks if individual frames look realistic given $\mathbf{y}_{a}$. $\mathcal{D}_{m}$ checks if a video contains realistic motion given $\mathbf{y}_{m}$. Note that either discriminator alone would be insufficient to achieve our goal: without $\mathcal{D}_{a}$ a generated video may have inconsistent visual appearance across frames, without $\mathcal{D}_{m}$ a generated video may not depict the motion we intend to hallucinate. 

The generator and the two discriminators form a conditional GAN~\cite{mirza-arxiv14}. This alone would be in sufficient to learn the role of conditioning variables unless a proper care is taken. If we follow the traditional training method~\cite{mirza-arxiv14}, the model may treat them as random noise. To ensure that the conditioning variables have intended influence on the data generation process, we employ a ranking network $\mathcal{R}$, which takes as input a triplet $(\mathbf{x}|_{\mathbf{y}}, \hat{\mathbf{x}}|_{\mathbf{y}}, \hat{\mathbf{x}}|_{\mathbf{y^{\prime}}})$ and forces $\mathbf{x}|_{\mathbf{y}}$ and $\hat{\mathbf{x}}|_{\mathbf{y}}$ to look more similar to each other than $\mathbf{x}|_{\mathbf{y}}$ and $\hat{\mathbf{x}}|_{\mathbf{y^{\prime}}}$, because in the latter pair, the conditions do not match ($\mathbf{y} \neq \mathbf{y}'$). 

In addition to the ranking constraint, we propose a novel conditioning scheme to put constraints on the learning objective with respect to the conditioning variables. We explain our learning strategy and the conditioning scheme in Section~\ref{subsec:learning_strategy}, and discuss model training in Section~\ref{subsec:model_training}.

\begin{figure*}
    \centering
    \vspace*{-2pt}
    \includegraphics[width=\linewidth]{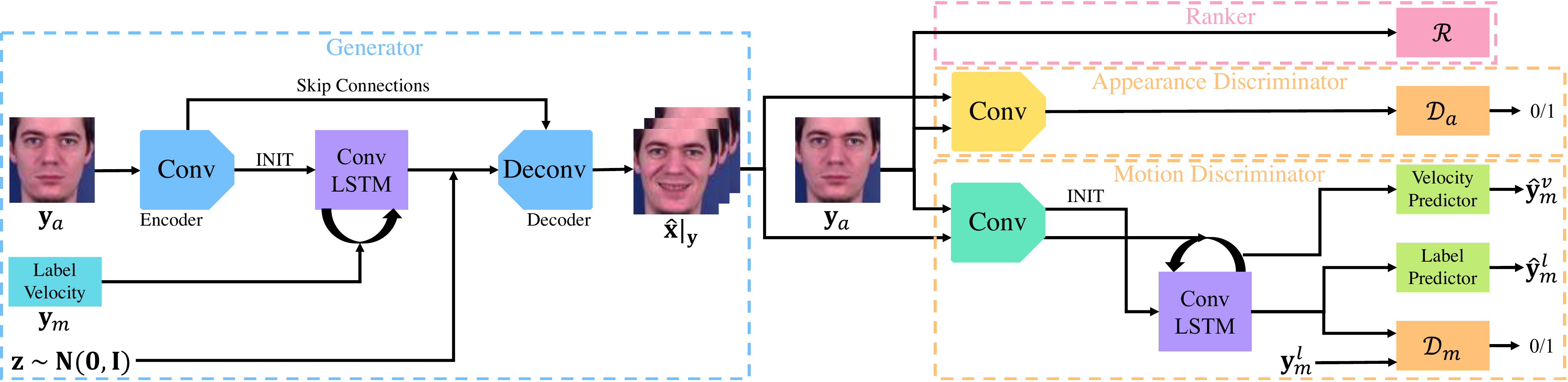}
    \vspace*{-1.6em}
    \caption{An overview of our AMC-GAN; we provide architecture and implementation details in the supplementary material.}
    \label{fig:architecture}
\end{figure*}

\subsection{Appearance and Motion Conditions}
\label{subsec:conditions}

The appearance condition $\mathbf{y}_{a}$ can be any high-level abstraction that encodes visual appearance; we use a single RGB image $\mathbf{y}_{a} \in \mathbb{R}^{64 \times 64 \times 3}$ (e.g., the first frame of a video).

The motion condition $\mathbf{y}_{m}$ can also be any high-level abstraction that encodes motion. We define it as $\mathbf{y}_{m} = \left[ \mathbf{y}^{l}_{m}, \mathbf{y}^{v}_{m} \right]$, where $\mathbf{y}^{l}_{m} \in \mathbb{R}^{c}$ is a motion category label encoded as a one-hot vector, and $\mathbf{y}^{v}_{m} \in \mathbb{R}^{(T-1) \times 2k}$ is the velocity of $k$ keypoints in 2D space detected from an image sequence of length $T$. We explain how we extract keypoints in Section~\ref{sec:experiments}. We repeat $\mathbf{y}^{l}_{m}$ $T-1$ times to obtain $\mathbf{y}_{m} \in \mathbb{R}^{(T-1) \times q}$, where $q=(c+2k)$. We set $T=32$ in all our experiments.

We assume $\mathbf{y}^{l}_{m}$ is known both during training and inference. However, we assume $\mathbf{y}^{v}_{m}$ is known \textit{only during training}; during inference, we randomly sample it from those training examples that share the same class $\mathbf{y}^{l}_{m}$ as the test example. 

\subsection{The Model}
\label{subsec:model}

We describe the four modules of our model (see Figure~\ref{fig:architecture}); implementation details are provided in the supplementary.

\textbf{Generator}: This has the encoder-decoder structure with a convLSTM~\cite{xingjian-nips15} in the middle. It takes as input the two conditioning variables $\mathbf{y}_{a}$ and $\mathbf{y}_{m}$, and a random noise vector $\mathbf{z} \in \mathbb{R}^{p}$ sampled from a normal distribution $\mathcal{N}(0,I)$. The output is a video $\hat{\mathbf{x}}|_{\mathbf{y}}$ generated frame-by-frame by unrolling the convLSTM for $T-1$~times.  

We use the encoder output to initialize the convLSTM. At each time step $t$, we provide the $t$-th slice of $\mathbf{y}_{m,t} \in \mathbb{R}^{q}$ to the convLSTM, and combine its output with the noise vector $\mathbf{z}$ and the encoder output. This becomes input to the image decoder. The noise vector $\mathbf{z}$, sampled once per video, introduces a certain degree of randomness to the decoder, helping the generator probe the distribution better~\cite{goodfellow-nips14}. We add a skip connection to create a direct path from the encoder to the decoder. This helps the model focus on learning changes in movement rather than full appearance and motion. We empirically found this to be crucial in producing high quality output.

\textbf{Appearance Discriminator}: This takes as input four images, an appearance condition $\mathbf{y}_{a}$ and three frames $\mathbf{x}_{t-1:t+1}$ from either a real or a generated video, and produces a scalar indicating whether the frame is real or fake. Note the conditional formulation with $\mathbf{y}_{a}$: This is crucial to ensure the appearance of generated frames is cohesive across time with the first frame, e.g., it should be facial movements that change over time, not its identity.

\textbf{Motion Discriminator}: This takes as input a video $\mathbf{x}=[\mathbf{x}_{1:T}]$ and the two conditions $\mathbf{y}_{a}$ and $\mathbf{y}^{l}_{m}$. It predicts three variables: a scalar indicating whether the video is real or fake, $\hat{y}_{m}^{l} \in \mathbb{R}^{c}$ representing motion categories, and $\hat{y}_{m}^{v} \in \mathbb{R}^{2k}$ representing the velocity of $k$ keypoints. The first is the adversarial discrimination task; we provide the motion category label $\mathbf{y}_{m}^{l}$ to perform class-conditional discrimination of the videos. The latter two are auxiliary tasks, similar to InfoGAN~\cite{chen-nips16} and BicycleGAN~\cite{zhu-nips17}, introduced to make our model more robust. We show the importance of the auxiliary tasks in Section~\ref{sec:experiments}.

\textbf{Perceptual Ranking:} This takes a triplet $(\mathbf{x}|_{\mathbf{y}}, \hat{\mathbf{x}}|_{\mathbf{y}}, \hat{\mathbf{x}}|_{\mathbf{y^{\prime}}})$ and outputs a scalar indicating the amount of violation for a constraint $d(\mathbf{x}|_{\mathbf{y}}, \hat{\mathbf{x}}|_{\mathbf{y}}) < d(\mathbf{x}|_{\mathbf{y}}, \hat{\mathbf{x}}|_{\mathbf{y}^{\prime}})$, where $d(\cdot,\cdot)$ is a function that computes a perceptual distance between two videos, and $\mathbf{y}' \neq \mathbf{y}$; we call $\mathbf{y}'$ a ``mismatched'' condition.

To compute the perceptual distance, we adapt the idea of the perceptual loss used in image style transfer~\cite{gatys-nips15,johnson-eccv16}, in which the distance is measured based on the feature representation at each layer of a pretrained CNN (e.g., VGG-16). In this work, we cannot simply use a pretrained CNN because of the conditioning variables; we instead use our own discriminator networks to compute them. Since we have two discriminators, we choose one based on a mismatched condition $\mathbf{y'}$, i.e., we use $\mathcal{D}_{a}$ when $\mathbf{y'} = [\mathbf{y}_{a'},\mathbf{y}_{m}]$ and $\mathcal{D}_{m}$ when $\mathbf{y'} = [\mathbf{y}_{a},\mathbf{y}_{m'}]$.

There are two ways to compute the perceptual distance: compare filter responses directly or the Gram matrices of the filter responses. The former encourages filter responses of a generated video to replicate, pixel-to-pixel, the ones of a training video. This is too restrictive for our purpose because we want our model to ``go beyond'' what exists in the training data; we want $\hat{\mathbf{x}}|_{\mathbf{y'}}$ to look realistic even if the given (video, condition) pair does not exist in the training set. The latter relaxes this restriction by encouraging filter responses to share similar correlation patterns between two videos. We take this latter approach in our work.

Let $G^{\mathcal{D}}_{j}(\cdot)$ be the Gram matrix computed at the $j$-th layer of a discriminator network $\mathcal{D}$. We define the distance function at the $j$-th layer of the network as
\begin{equation}
\label{eq:distance_function}
d_{j}(\mathbf{x}|_{\mathbf{y}}, \hat{\mathbf{x}}|_{\mathbf{y}}) = \left\| G^{\mathcal{D}}_j(\mathbf{x}|_{\mathbf{y}}) - G^{\mathcal{D}}_j(\hat{\mathbf{x}}|_{\mathbf{y}}) \right\|_F
\end{equation}
where $\left\|\cdot\right\|_F$ is the Frobenius norm. To compute the Gram matrix, we reshape the output of the $j$-th layer from a discriminator network  to be the size of $N_j \times M_j$, where $N_j = T_j \times C_j$ (sequence length $\times$ number of channels) and $M_j = H_j \times W_j$ (height $\times$ width). Denoting this reshaped matrix by $\omega_{j}(\mathbf{x})$, the Gram matrix is
\begin{equation}
\label{eq:gram_matrix}
G^{\mathcal{D}}_j(\mathbf{x}) = \omega_{j}(\mathbf{x})^{\top} \omega_{j}(\mathbf{x})~/~{N}_{j}{M}_{j}.
\end{equation}
Finally, we employ triplet ranking~\cite{wang-cvpr14,schroff-cvpr15} to measure the amount of violation, using $\mathbf{x}|_{\mathbf{y}}$ as an anchor point and $\hat{\mathbf{x}}|_{\mathbf{y}}$ and $\hat{\mathbf{x}}|_{\mathbf{y}^{\prime}}$ as positive and negative samples, respectively. Specifically, we use the hinge loss form to quantify the amount of violation:
\begin{equation}
\label{eq:tripletloss_definition}
\mathcal{R}(\mathbf{x}|_{\mathbf{y}}, \hat{\mathbf{x}}|_{\mathbf{y}}, \hat{\mathbf{x}}|_{\mathbf{y^{\prime}}}) = \sum\nolimits_{j} \max \left( 0, \rho - d_{j}^{-} + d_{j}^{+} \right)
\end{equation}
where $\rho$ determines the margin between positive and negative pairs (we set $\rho$ as 0.01 for $\mathcal{D}_{a}$ and 0.001 for $\mathcal{D}_{m}$), $d_{j}^{+} = d_{j}(\mathbf{x}|_{\mathbf{y}}, \hat{\mathbf{x}}|_{\mathbf{y}})$ and $d_{j}^{-} = d_{j}(\mathbf{x}|_{\mathbf{y}}, \hat{\mathbf{x}}|_{\mathbf{y'}})$, and $j=[1,2]$.

\subsection{Learning Strategy}
\label{subsec:learning_strategy}

We specify three constraints on the behavior of our model to help it learn the data distribution effectively:
\begin{compactitem}
\item \textbf{C1}: If we take one of the training samples $\mathbf{x}|_{\mathbf{y}}$ and pair it with a different condition, i.e., $(\mathbf{x}|_{\mathbf{y}}, \mathbf{y'})$, our discriminators should be able to tell the pair is fake.
\item \textbf{C2}: Regardless of the input condition, videos produced by the generator should be able to fool the discriminators into believing that $\hat{\mathbf{x}}|_{\mathbf{y}}$ and $\hat{\mathbf{x}}|_{\mathbf{y'}}$ are real.
\item \textbf{C3}: The pair ($\mathbf{x}|_{\mathbf{y}}, \hat{\mathbf{x}}|_{\mathbf{y}}$) should look more similar to each other than the pair ($\mathbf{x}|_{\mathbf{y}}, \hat{\mathbf{x}}|_{\mathbf{y'}}$) because the former shares the same condition (in the latter, $\mathbf{y}\neq\mathbf{y}'$).
\end{compactitem}

\begin{table}[t]
\small
\centering
\begin{tabular}{|c|c||c|}
\hline
Appearance & Motion & Output \\ \hline
$\mathbf{y}_{a}$ & $\mathbf{y}_{m}$ & $\hat{\mathbf{x}}|_{\mathbf{y}_{a},\mathbf{y}_{m}} = \mathcal{G}(\mathbf{z}~|~\mathbf{y}_{a}, \mathbf{y}_{m})$ \\ \hline
$\mathbf{y}_{a^{\prime}}$ & $\mathbf{y}_{m}$ & $\hat{\mathbf{x}}|_{\mathbf{y}_{a^{\prime}},\mathbf{y}_{m}} = \mathcal{G}(\mathbf{z}~|~\mathbf{y}_{a^{\prime}}, \mathbf{y}_{m})$ \\ \hline
$\mathbf{y}_{a}$ & $\mathbf{y}_{m^{\prime}}$ & $\hat{\mathbf{x}}|_{\mathbf{y}_{a},\mathbf{y}_{m^{\prime}}} = \mathcal{G}(\mathbf{z}~|~\mathbf{y}_{a}, \mathbf{y}_{m^{\prime}})$ \\ \hline
\end{tabular}
\vspace{-.5em}
\caption{Three conditions used in the generator network.}
\label{tab:condition_pair_g} 

\vspace{1em}

\begin{tabular}{|cccc|cccc|}
\hline
\multicolumn{4}{|c|}{$\mathcal{D}_{a}$} & \multicolumn{4}{c|}{$\mathcal{D}_{m}$} \\ \hline
$\mathbf{x}$    & $\mathbf{y}$ & $\mathcal{G}$ & $\mathcal{D}$       & $\mathbf{x}$ & $\mathbf{y}$ & $\mathcal{G}$ & $\mathcal{D}$ \\ \hline
$\mathbf{x}|_{\mathbf{y}_{a},\mathbf{y}_{m}}$ & $\mathbf{y}_{a}$            & -      & \cmark & $\mathbf{x}|_{\mathbf{y}_{a},\mathbf{y}_{m}}$ &  $\mathbf{y}_{m}$               & -      & \cmark \\ 
$\mathbf{x}|_{\mathbf{y}_{a},\mathbf{y}_{m}}$ &  $\mathbf{y}_{a^{\prime}}$ & -   & \xmark & $\mathbf{x}|_{\mathbf{y}_{a},\mathbf{y}_{m}}$ & $\mathbf{y}_{m^{\prime}}$   & -      & \xmark \\ \hline
$\hat{\mathbf{x}}|_{\mathbf{y}_{a},\mathbf{y}_{m}}$ & $\mathbf{y}_{a}$          & \cmark & \xmark & $\hat{\mathbf{x}}|_{\mathbf{y}_{a},\mathbf{y}_{m}}$ & $\mathbf{y}_{m}$              & \cmark & \xmark \\ 
$\hat{\mathbf{x}}|_{\mathbf{y}_{a^{\prime}},\mathbf{y}_{m}}$ & $\mathbf{y}_{a^{\prime}}$ & \cmark & \xmark & $\hat{\mathbf{x}}|_{\mathbf{y}_{a},\mathbf{y}_{m^{\prime}}}$ & $\mathbf{y}_{m^{\prime}}$   & \cmark & \xmark \\ \hline
\end{tabular}
\vspace{-.5em}
\caption{Four conditions used in each discriminator network, with labels for the generator and discriminators: real (\cmark) or fake (\xmark).}
\vspace{-1em}
\label{tab:condition_pair_d} 
\end{table}

\textbf{Conditioning Scheme:} We provide three conditions to the generator, listed in Table~\ref{tab:condition_pair_g}. The first contains the original condition ($\mathbf{y}_{a}$, $\mathbf{y}_{m}$) matched with a training video $\mathbf{x}|_{\mathbf{y}_{a},\mathbf{y}_{m}}$. The other two pairs contain mismatched information on either variable. We select the mismatched condition by randomly selecting another condition from the training set. Note that we do not feed the pair ($\mathbf{y}_{a^{\prime}}$, $\mathbf{y}_{m^{\prime}}$) to the generator as it is equivalent to one of the other three combinations.

We provide four conditions to each discriminator, listed in Table~\ref{tab:condition_pair_d}. The first and the third rows are identical to conditional GAN~\cite{mirza-arxiv14}. Training our model with just these two conditions may make our model treat the conditioning variables as random noise in the worst case. This is because there is no constraint on the expected behavior of the conditioning variables on the generation process, other than just having the end results look realistic. 

We provide $(\mathbf{x}|_{\mathbf{y}}, \mathbf{y'})$ to the discriminators (the second row in Table~\ref{tab:condition_pair_d}) and have them identify it as fake; this enforces the constraint \textbf{C1}. Note that there is no gradient flow back to the generator because it has no control over $\mathbf{x}|_{\mathbf{y}}$. A similar idea was used by \cite{reed-icml16}, where they used a mismatched sentence for the text-to-image synthesis task. We provide $(\hat{\mathbf{x}}|_{\mathbf{y'}}, \mathbf{y'})$ to the discriminators (the fourth row) to enforce the constraint \textbf{C2}. With this, the generator needs to work harder to fool the discriminators because this condition does not exist in the training set. We do not include ($\hat{\mathbf{x}}|_{\mathbf{y}}$, $\mathbf{y'}$) and ($\hat{\mathbf{x}}|_{\mathbf{y'}}$, $\mathbf{y}$) because the conditions used in generator do not match with the conditions provided to the discriminator.

\subsection{Model Training}
\label{subsec:model_training}

Our learning objective is to solve the min-max game: 
\begin{align}
\label{eq:loss_definition}
\nonumber \min_{\theta_{\mathcal{G}}} \max_{\theta_{\mathcal{D}}} ~& \mathcal{L}_\mathit{gan}(\theta_{\mathcal{G}},\theta_{\mathcal{D}}) + \mathcal{L}_\mathit{rank}(\theta_{\mathcal{G}}) \\
& + \mathcal{L}_{\mathcal{D}_\mathit{aux}}(\theta_{\mathcal{D}}) + \mathcal{L}_{\mathcal{G}_\mathit{aux}}(\theta_{\mathcal{G}})
\end{align} 
where each $\mathcal{L}_{\{\cdot\}}$ has its own loss weight to balance the influence of it (see supplementary). The first term follows the conditional GAN objective~\cite{mirza-arxiv14}:
\begin{align}
\label{eq:ganloss_definition}
\nonumber \mathcal{L}_\mathit{gan}(\theta_{\mathcal{G}},\theta_{\mathcal{D}})~&=~\mathbb{E}_{\mathbf{x} \sim p_{\text{data}}(\mathbf{x})}[\log \mathcal{D}(\mathbf{x}|\mathbf{y})] ~~~~~~~~~~~~~~~~~~~~~~~~~\\
&+~\mathbb{E}_{\mathbf{z} \sim p_{z}(\mathbf{z})}[\log (1 - \mathcal{D}(\mathcal{G}(\mathbf{z}|{\mathbf{y}})|\mathbf{y}))]
\end{align}
where we collapsed $\mathcal{D}_{a}$ and $\mathcal{D}_{m}$ into $\mathcal{D}$ for brevity. We use the cross entropy loss for the real/fake discriminators. The second term is our perceptual ranking loss (see Eqn.~\eqref{eq:tripletloss_definition})
\begin{equation}
\label{eq:rankloss_definition}
\mathcal{L}_\mathit{rank}(\theta_{\mathcal{G}}) = \mathcal{R}(\mathbf{x}|_{\mathbf{y}}, \hat{\mathbf{x}}|_{\mathbf{y}}, \hat{\mathbf{x}}|_{\mathbf{y^{\prime}}}).
\end{equation}
The two terms play complementary roles during training: The first encourages the solution to satisfy \textbf{C1} and \textbf{C2}, while the second encourages the solution to satisfy \textbf{C3}. 

The third term is introduced to increase the power of our motion discriminator: 
\begin{equation}
\label{eq:auxliary_loss}
\mathcal{L}_{\mathcal{D}_\mathit{aux}} = \mathcal{L}_\mathit{CE}(\mathbf{y}_{m}^{l}, \hat{\mathbf{y}}_{m}^{l}) + \mathcal{L}_\mathit{MSE}(\mathbf{y}_{m}^{v},\hat{\mathbf{y}}_{m}^{v})
\end{equation}
where the first term is the cross entropy loss for predicting motion category labels, and the second is the mean square error loss for predicting the velocity of keypoints. 

The fourth term is introduced to increase the power of the generator, and is similar to the reconstruction loss widely used in video prediction~\cite{mathieu-iclr16},
\begin{equation}
\label{eq:recon_loss}
\mathcal{L}_{\mathcal{G}_\mathit{aux}} = \|\mathbf{x}|_{\mathbf{y}} - \hat{\mathbf{x}}|_{\mathbf{y}}\|_{1} + \sum\nolimits_{j} d_{j}(\mathbf{x}|_{\mathbf{y}}, \hat{\mathbf{x}}|_{\mathbf{y}}).
\end{equation}
\begin{algorithm}[tp]
    \begin{algorithmic}[1] 
    \begin{small}
        \STATE {\bfseries Input:} Dataset $\{\mathbf{x}|_\mathbf{y}\}$, conditions $\mathbf{y}$ and $\mathbf{y}^{\prime}$, step size $\eta$
        \FOR{each step}
            \STATE $\mathbf{z} \sim \mathcal{N}(0,I)$
            \STATE $\mathbf{\hat{x}}|_{\mathbf{y}} \gets \mathcal{G}(\mathbf{z}|\mathbf{y}),~~~~~~\mathbf{\hat{x}}|_{\mathbf{y}^{\prime}} \gets \mathcal{G}(\mathbf{z}|\mathbf{y}^{\prime})$
            \STATE $({s}_{r}, {v}_{r}, {l}_{r}) \gets \mathcal{D}(\mathbf{x}|_\mathbf{y}, \mathbf{y})$,  $({s}_{m}, {v}_{m}, {l}_{m}) \gets \mathcal{D}(\mathbf{x}|_\mathbf{y}, \mathbf{y}^{\prime})$
            \STATE[] $({s}_{f}, {v}_{f}, {l}_{f}) \gets \mathcal{D}(\mathbf{\hat{x}}|_\mathbf{y}, \mathbf{y})$, $({s}_{f^{\prime}}, {v}_{f^{\prime}}, {l}_{f^{\prime}}) \gets \mathcal{D}(\mathbf{\hat{x}}|_\mathbf{y^{\prime}}, \mathbf{y}^{\prime})$
            
            \STATE $\mathcal{L}_{\mathcal{D}} \gets \log({s}_{r}) + 0.5[\log(1-{s}_{m})$
            \STATE[] $~~~~~~~~~~~~~~~~~~~+ 0.5(\log(1-{s}_{f}) + \log(1-{s}_{f^{\prime}}))]$
            \STATE $\mathcal{L}_{\mathcal{D}_\mathit{aux}} \gets \left[\left\| y^{v}_{m} - {v}_{r} \right\|^{2}_{2} + \left\| y^{v}_{m} - {v}_{f} \right\|^{2}_{2} + \left\| {{y}^{\prime}}^{v}_{m} - {v}_{f^{\prime}} \right\|^{2}_{2} \right] $ 
            \STATE[] $~~~~~~~~~~- {\scriptstyle\sum}_{i} y^{l}_{m,i} \left[\log y({l}_{r,i}) + \log y({l}_{f,i}) + \log y({l}_{f^{\prime},i})\right]$
            \STATE ${\theta}_{\mathcal{D}} \gets {\theta}_{\mathcal{D}} + \eta\frac{\partial(\mathcal{L}_{\mathcal{D}} -  \mathcal{L}_{\mathcal{D}_\mathit{aux}})}{\partial{\theta}_{\mathcal{D}}}$
            \STATE $d^{+}_{j} = \left\| G^{\mathcal{D}}_j(\mathbf{x}|_\mathbf{y}) - G^{\mathcal{D}}_j(\mathbf{\hat{x}}|_\mathbf{y}) \right\|_{F}$ for $j=1,2$
            \STATE $d^{-}_{j} = \left\| G^{\mathcal{D}}_j(\mathbf{x}|_\mathbf{y}) - G^{\mathcal{D}}_j(\mathbf{\hat{x}}|_\mathbf{y'}) \right\|_{F}$ for $j=1,2$
            \STATE $\mathcal{L}_{\mathcal{G}} \gets \log({s}_{f}) + \log({s}_{f^{\prime}})$
            \STATE $\mathcal{L}_{\mathcal{G}_\mathit{aux}} \gets  {\scriptstyle\sum}^{T}_{t=2} \left\|\mathbf{x}_t|_\mathbf{y} - \mathbf{\hat{x}}_t|_\mathbf{y}\right\|_1 + {\scriptstyle\sum}^{2}_{j=1} d^{+}_{j}$
			\STATE $\mathcal{L}_\mathit{rank} \gets  {\scriptstyle\sum}^{2}_{j=1} \max (0, \rho - d^{-}_{j} + d^{+}_{j}) $
            \STATE ${\theta}_{\mathcal{G}} \gets {\theta}_{\mathcal{G}} + \eta\frac{\partial(\mathcal{L}_{\mathcal{G}}~-~\mathcal{L}_{{\mathcal{G}_\mathit{aux}}}~-~\mathcal{L}_\mathit{rank})}{\partial{\theta}_{\mathcal{G}}}$
        \ENDFOR
    \end{small}
    \end{algorithmic}
    \caption{AMC-GAN Training Algorithm \label{alg:training}}
\end{algorithm}
Algorithm~\ref{alg:training} summarizes how we train our model. We solve the bi-level optimization problem where we alternate between solving for $\theta_{\mathcal{D}}$ with respect to the optimum of $\theta_{\mathcal{G}}$ and vice versa. We train the discriminator networks based on a mini-batch containing a mix of the four cases listed in Table~\ref{tab:condition_pair_d}. We put different weights to each of the four cases (Line 7), as suggested by \cite{reed-icml16}. The generator is trained on a mini-batch of the three cases listed in Table~\ref{tab:condition_pair_g}. We use the ADAM optimizer~\cite{kingma-iclr15} with learning rate 2e-4. For the cross entropy losses, we adopt the label smoothing trick~\cite{salimans-nips16} with a weight decay of 1e-5 per mini-batch~\cite{arjovsky-iclr17}.

\section{Experiments}
\label{sec:experiments}

We evaluate our approach on the MUG facial expression dataset~\cite{aifanti-wiamis10} and the NATOPS human action dataset~\cite{song-fg11}. The MUG dataset contains 931 video clips performing six basic emotions~\cite{ekman-cogemo1992} (anger, disgust, fear, happy, sad, surprise). We preprocess it so that each video has 32 frames with 64 $\times$ 64 pixels (see supplementary for details). We use 11 facial landmark locations (2, 9, 16, 20, 25, 38, 42, 45, 47, 52, 58th) as keypoints for each frame, detected using the OpenFace toolkit~\cite{baltruvsaitis-wacv16}. The NATOPS dataset contains 9,600 video clips performing 24 action categories. We crop the video to 180 $\times$ 180 pixels with the chest at the center position and rescale it to 64 $\times$ 64 pixels. We use 9 joint locations (head, chest, naval, L/R-shoulders, L/R-elbows, L/R-wrists) as keypoints for each frame, provided by the dataset.

\subsection{Quantitative Evaluation}
\label{subsec:quantative_evaluation}

\textbf{Methodology:} We design a $c$-way motion classifier using a 3D CNN~\cite{tran-iccv15} that predicts the motion label $\mathbf{y}^{l}_{m}$ from a video (see the supplementary for the architecture). To prevent the classifier from predicting the label simply by seeing the input frame(s), we only use the last 28 generated frames as input. We train the classifier on real training data, using roughly 10\% for validation, and test it on generated videos from different methods. 

We compare our method with recent approaches in video prediction: CDNA~\cite{finn-nips16}, Adv+GDL with $\ell_1$ loss~\cite{mathieu-iclr16}, and MCnet~\cite{villegas-iclr17}. For CDNA, we provide $\mathbf{y}_{a}$ as input image and $\mathbf{y}_{m}$ as the ``action \& state'' variable. We use 10 masks suggested in their work, and disable teacher forcing for fair comparison with other methods. Following the original implementations for Adv+GDL and MCnet, we provide as input the first \textit{four} consecutive frames, but no $\mathbf{y}_{m}$. 
We also perform ablative analyses by eliminating various components of our method; we explain various settings as we discuss the results.

\begin{table}[t]
\small
\centering
\begin{tabular}{|c|c||c|c|}
\hline
\multicolumn{2}{|c||}{Method} & MUG & NATOPS \\\hline\hline
\multicolumn{2}{|c||}{Random} & 16.67 & 4.17 \\\hline
\multicolumn{2}{|c||}{CDNA$\dagger$~\cite{finn-nips16}} & 35.38 & 6.80  \\ 
\multicolumn{2}{|c||}{Adv+GDL$\ddagger$~\cite{mathieu-iclr16}} & 40.47 & 8.45 \\ 
\multicolumn{2}{|c||}{MCnet$\ddagger$~\cite{villegas-iclr17}} & 43.22 & 12.76  \\ \hline 
\multicolumn{2}{|c||}{\textbf{AMC-GAN}$\dagger$ (ours)} &  \textbf{99.79} & \textbf{91.12} \\\hline 
\end{tabular}
\caption{Video classification results (accuracy). Models with $\dagger$ learn from a single input frame, while $\ddagger$ use four input frames.}
\vspace{-1em}
\label{tab:quantitative_evaluation} 
\end{table}

\textbf{Results:} Table~\ref{tab:quantitative_evaluation} shows the results. 
We notice that the CDNA performs worse than the other methods. This is expected because it predicts future frames by combining multiple frames via masking, each generated by shifting the entire pixels of the previous frame in a certain direction. Our datasets contain complex object deformations that cannot be synthesized simply by shifting pixels. Because our network predicts pixel values directly, we achieve better results on more naturalistic videos. Both Adv+GDL and MCnet outperforms CDNA but not ours. We believe this is because both models learn to extrapolate past observations into the future. Therefore, if the input (four consecutive frames) do not provide enough motion information, as is true in our case (most videos start with ``neutral'' faces and body poses), extrapolation fails to predict future frames. Lastly, our model outperforms all the baselines by significant margins. It is because our model is \textit{optimized} to generate videos that fool the motion discriminator with $\mathcal{L}_{\mathcal{D}_\mathit{aux}}$, which guides our model to preserve well the property of the motion condition $\mathbf{y}_{m}$.

To verify whether our model successfully generates videos with the \textit{correct} motion information provided as input, we run a similar experiment on the MUG dataset with only keypoints extracted from the generated output. For this, we use the OpenFace toolkit~\cite{baltruvsaitis-wacv16} to extract 68 facial landmarks from the predicted output and render them with a Gaussian blur on a 2D grid to produce grayscale images. This is then fed into a $c$-way 2D CNN classifier (details in the supplementary). The results confirm that our method produces videos with the most accurate keypoint trajectories, with an accuracy of 70.34\%, compared to CDNA (23.52\%), Adv+GDL (28.81\%), MCnet (35.38\%).

\begin{table}[t]
\small
\centering
\begin{tabular}{|c|c||c|c|}
\hline
No $\mathbf{y}_{a}$ 	& 4.75					& No $\mathcal{D}_{a}$ 							& 85.97 \\ 
No $\mathbf{y}_{m}$ 	& 7.62 					& No $\mathcal{D}_{m}$ 							& 79.23 \\\hline 
No $\mathbf{y}$, $\mathbf{y}'$ 		& 5.52		& No $\mathcal{L}_{\mathcal{D}_\mathit{aux}}$ 	& 81.05 \\ 
No $\mathbf{y}'$ 		& 86.80					& No $\mathcal{L}_{\mathcal{G}_\mathit{aux}}$ 	& 88.29 \\\cline{1-2} 
\textbf{Ours} 			& \textbf{91.12} 		& No $\mathcal{L}_\mathit{rank}$ 				& 90.83 \\\hline 
\end{tabular}
\caption{Ablation study results on the NATOPS dataset (accuracy).}
\vspace{-1em}
\label{tab:abalation} 
\end{table}

For an ablation study, we remove input conditions and their corresponding discriminators to measure the relative importance of appearance and motion. Not surprisingly, removing either $\mathbf{y}_{a}$ or $\mathbf{y}_{m}$ signifncantly drops the performance. Similarly, having no $\mathbf{y}$ and $\mathbf{y}'$ (i.e., produce videos solely based on random noise $\mathbf{z}$) results in poor performance. Finally, we remove the mismatched conditions $\mathbf{y}'$ from our conditioning scheme, i.e., we use only the first row of Table~\ref{tab:condition_pair_g} and the first and third rows of Table~\ref{tab:condition_pair_d}; this is similar to the standard conditional GAN~\cite{mirza-arxiv14}. We can see a performance drop. This is because our model ends up treating the conditioning variables alike to random noise; without contradicting conditions, the discriminators have no chance of learning to discriminate different conditions.

Removing $\mathcal{D}_{m}$ shows significant drop in performance, which is expected because without motion constraints the model is incentivized to produce what appears as a static image (repeat $\mathbf{y}_{a}$ to make the appearance realistic). Removing $\mathcal{D}_{a}$ also decreases the performance, but not as much as removing $\mathcal{D}_{m}$. This is however deceptive: videos look visually implausible and appear to be adversarial examples (the faces under ``No $\mathcal{D}_{a}$'' in Fig. 3). This shows the importance of enforcing constraints on visual appearance: without it, the model ``over-optimizes'' for the motion constraint. 

Finally, we remove loss terms from our objective Eqn.~\eqref{eq:loss_definition}. Removing $\mathcal{L}_{\mathcal{D}_\mathit{aux}}$ significantly deteriorates our model. This shows its effectiveness in enhancing the power of the motion discriminator; without it, similar to removing $\mathcal{D}_{m}$, the model is less constrained to predict realistic motion. Removing $\mathcal{L}_{\mathcal{G}_\mathit{aux}}$ decreases the performance moderately.
Visual inspection of the generated videos revealed that this has a similar effect to removing $\mathcal{D}_{a}$ or $\mathbf{y}_{a}$; the model over-optimizes for the motion constraint. This is consistent with the literature~\cite{mathieu-iclr16,shrivastava-cvpr17} that shows the effectiveness of the reconstruction constraint. Removing $\mathcal{L}_\mathit{rank}$ hurts the performance only marginally; however, we found that the ranking loss improves visual quality and leads to faster model convergence.

\subsection{Qualitative Results}
\label{subsec:qualitative_evaluation}

\begin{table}[t]
\small
\centering
\begin{tabular}{c||c|c}
Preference & MUG & NATOPS \\\hline
Prefers ours over CDNA & 77.2\% & 97.1\% \\
Prefers ours over Adv+GDL & 80.4\% & 91.4\% \\
Prefers ours over MCnet & 72.4\% & 98.2\% \\\hline
Prefers ours over the ground truth & 13.9\% & 5.3\% \\
\end{tabular}
\vspace{-0.5em}
\caption{Human subjective preference results.}
\vspace{-1em}
\label{tab:qualitative_evaluation} 
\end{table}

\textbf{Methodology:} We adapt the evaluation protocol from \cite{vondrick-nips16} and ask humans to specify their subjective preference when given a pair of videos generated using different methods under the same condition. We randomly chose 100 videos from the test split of each dataset and created 400 video pairs. We recruited 10 participants for this study; each rated all the pairs from each dataset.

\textbf{Results:} Table~\ref{tab:qualitative_evaluation} shows the results when the participants are provided with motion category information along with videos. This ensures that their decision takes into account both \textit{appearance} and \textit{motion}; without the category label, their decision is purely based on appearance. Our participants significantly preferred ours over the baselines. Notably, for the NATOPS dataset, more than 90\% of the participants voted for ours. This is because the dataset is more challenging with more categories (24 actions vs. 6 emotions); a model must generate plausibly looking videos (appearance) with distinct movements across categories (motion), which is more challenging with more categories.   

To evaluate the quality of the generated videos in terms of appearance and motion separately, we designed another experiment with two tasks: We give the participants the same preference task but without motion category information. Subsequently, we ask them to identify which of the 7 facial expressions (neutral and 6 emotions) is depicted in each generated video. These two tasks focus on appearance and motion, respectively. Our participants preferred ours over CDNA (80.8\%), Adv+GDL (86.4\%), MCnet (55\%), and the ground truth (5\%). The MCnet approach was a close match, showing that videos generated by ours and MCnet have a similar quality in terms of appearance. However, results from the second task showed that none of the three baselines successfully produced videos with distinct motion patterns: The human classification accuracy was: Ours 66\%, CDNA 7\%, Adv+GDL 3\%, MCnet 7\%, GT: 77\%. This suggests that MCnet, while producing visually plausible output, fails to produce videos with intended motion.

\begin{figure*}
    \centering
    \includegraphics[width=\linewidth]{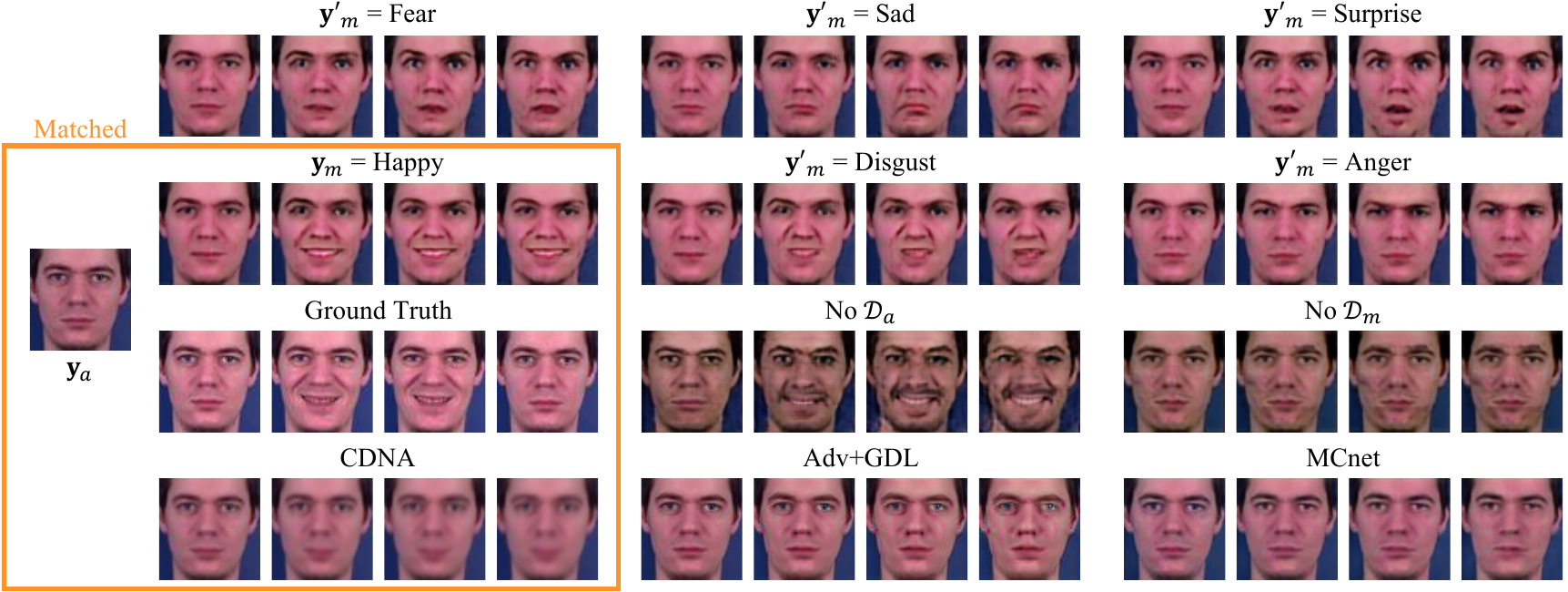}
        
    \vspace{12pt}
    
    \includegraphics[width=\linewidth]{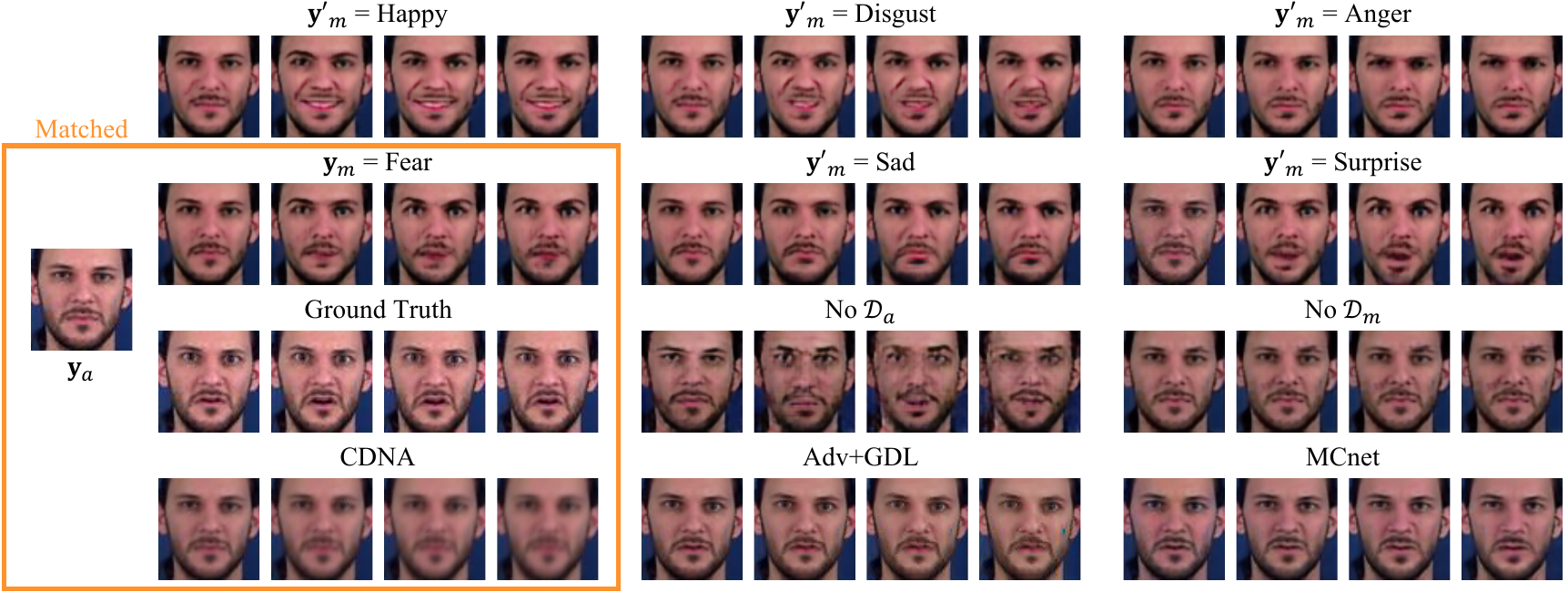}
    
    \vspace{12pt}
    
    \includegraphics[width=\linewidth]{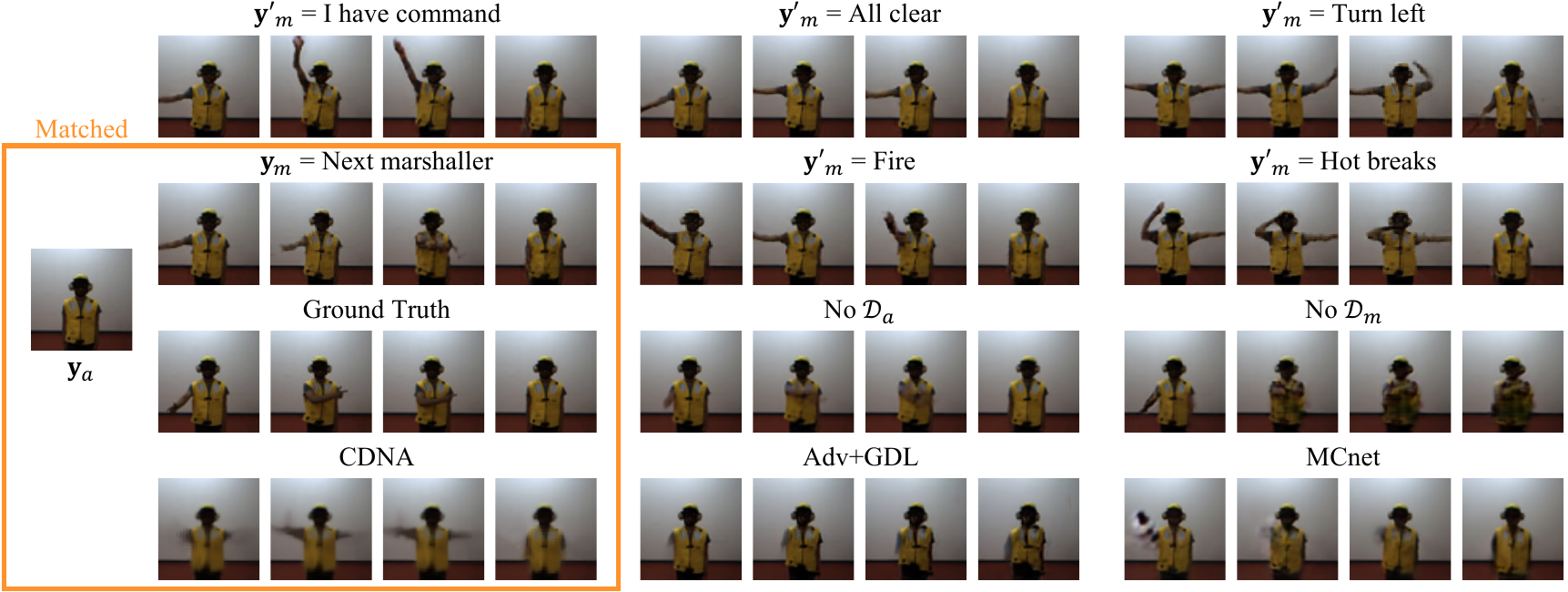}
    \caption{Video prediction results; $\mathbf{y}_{a}$ is the appearance condition (input frame), $\mathbf{y}_{m}$ is the motion condition, four frames under each method are generated results (8/16/24/32th frames). We show our approach generating different videos using matched ($\mathbf{y}_{m}$) and mismatched ($\mathbf{y}'_{m}$) motion conditions. We also show our ablation results (No $\mathcal{D}_{a}$/$\mathcal{D}_{m}$) and the baseline results. See the text for discussion. To see more results in video format, we invite the readers to visit our project page at \href{http://vision.snu.ac.kr/projects/amc-gan}{\color{magenta}http://vision.snu.ac.kr/projects/amc-gan}.}
    \label{fig:qualitative_results}
\end{figure*}

Figure~\ref{fig:qualitative_results} shows generated videos. Our method produces noticeably sharper frames and manifests more distinct/correct motion patterns than the baselines. Most importantly, the results show that we can \textit{manipulate} the future frames by changing the motion condition; notice how the same input frame $\mathbf{y}_{a}$ turns into different videos. The results also show the importance of appearance and motion discriminators. Removing $\mathcal{D}_{a}$ deteriorates the visual realism in the output: While the results still manifest the intended motion (``happy'' in the first set of examples), the generated frames look visually implausible (the face identity changes over time). Removing $\mathcal{D}_{m}$ produces what appears as a static video. 

The CDNA produces blurry frames without clear motion, despite the fact that it receives the same $\mathbf{y}_{a}$ and $\mathbf{y}_{m}$ as our model. MCnet and Adv+GDL receive four-frame input frames, which provide appearance and motion information. While the results are sharper than the CDNA, we see motion patterns are not as distinct/correct as ours (they look almost stationary), due to future uncertainty caused by too little motion information exist in the input. This suggests that the ``learning to extrapolate'' approaches do not successfully address the ill-conditioning issue in video prediction.

The results from our quantitative and qualitative experiments highlight the advantage of our approach: Disentangling appearance from motion in the input space and learning dynamic visual representation using our method produces higher-fidelity videos than the compared methods, which suggests that our method learns video representation more precisely than the baselines. 

\section{Conclusion}
\label{sec:conclusion}

We presented an AMC-GAN to address the future uncertainty issue in video prediction. The decomposition of appearance and motion conditions enabled us to design a novel conditioning scheme, which puts constraints on the behavior of videos generated under different conditions. We empirically demonstrated that our method produces sharp videos with the content expected by input conditions better than alternative solutions.

\appendix
\section*{Appendix}
\setcounter{table}{0}
\setcounter{figure}{0}
\renewcommand{\thetable}{\Alph{table}}
\renewcommand{\thefigure}{\Alph{figure}}

\section{Architecture Details (Section 3.2)}
\label{sec:model}

We provide architecture details of our AMC-GAN intruduced in the main paper (Section 3.2).

\begin{figure}[t]
    \centering
    \includegraphics[width=\linewidth]{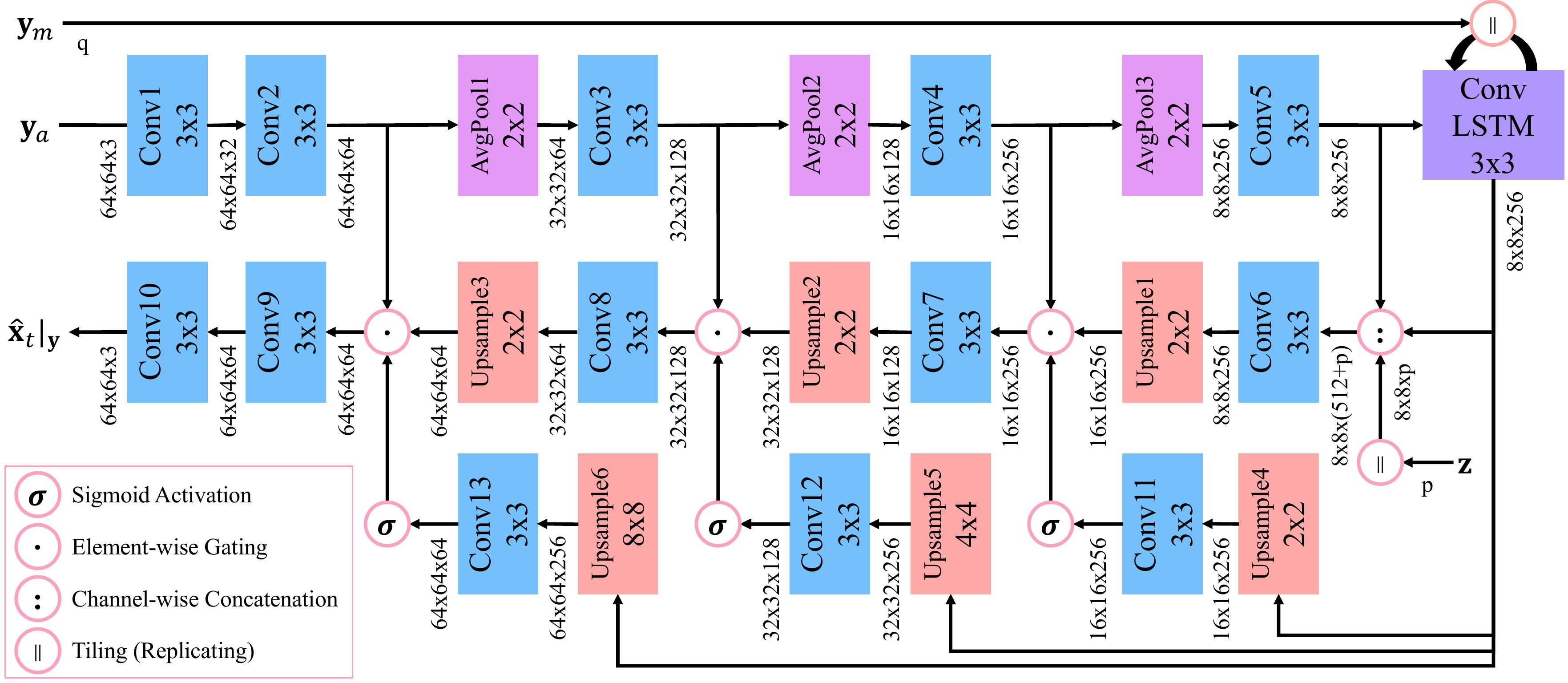}
    \caption{Generator $\mathcal{G}$ network architecture (used in Figure~2 of the main paper).}
    \label{fig:generator}
\end{figure}

\subsection{Generator Network (Figure~\ref{fig:generator})}

It takes a random noise vector $\mathbf{z} \in \mathbb{R}^{p}$ sampled from a normal distribution $\mathcal{N}(0,I)$, and the two conditioning variables $\mathbf{y}_{a}$ and $\mathbf{y}_{m}$ as input; we set $p=96$ for MUG and $128$ for NATOPS. The output is a video $\hat{\mathbf{x}}|_{\mathbf{y}}$, generated frame-by-frame by unrolling a convolutional LSTM (convLSTM)~\cite{xingjian-nips15} and an image decoder network $T-1$ times.  

We encode $\mathbf{y}_{a}$ using five convolutional layers: \texttt{conv2d(32)} -- \texttt{leakyReLU} -- \texttt{conv2d(64)} -- \texttt{BN} -- \texttt{leakyReLU} -- \texttt{pool} -- \texttt{conv2d(128)} -- \texttt{BN} -- \texttt{leakyReLU} -- \texttt{pool} -- \texttt{conv2d(256)} -- \texttt{BN} -- \texttt{leakyReLU} -- \texttt{pool} -- \texttt{conv2d(256)} -- \texttt{BN} -- \texttt{leakyReLU}, where \texttt{conv2d($k$)} is a 2D convolutional layer with $k$ filters of $3 \times 3$ kernel with stride 1, \texttt{pool} is average pooling on $2 \times 2$ region with stride 2, and \texttt{BN} is batch normalization~\cite{ioffe-icml2015}. The output is an embedding $\phi(\mathbf{y}_{a})$ of size $8 \times 8 \times 256$. 

We unroll the convLSTM~\cite{xingjian-nips15} for $T-1$ time steps to produce the output video. The convLSTM has $256$ filters of $3 \times 3$ kernel with stride 1. 
We initialize its states using $\phi(\mathbf{y}_{a})$.

At each $t$-th time step, we pass the motion condition $\mathbf{y}_{m,t} \in \mathbb{R}^{q}$ to a fully connected layer with a gated operation. That is, we compute the gate value $t = sigmoid(fc(q)) \in \mathbb{R}^{1}$ and obtain $t * fc(q) + (1-t) * q$. Then, we spatially tile it to form a $8 \times 8 \times q$ tensor, which is the input to the convLSTM. In our experiments, $q=28$ for the MUG dataset (by concatenating 11 of 2D facial landmarks and an one-hot vector of 6 emotion class) and $q=42$ for the NATOPS dataset (by concatenating 9 of 2D body joints and an one-hot vector of 24 motion class).

We add a skip connection to create a direct path from $\phi(\mathbf{y}_{a})$ to output of the convLSTM via channel-wise concatenation. We then apply the spatial tiling to the random noise vector $\mathbf{z} \in \mathbb{R}^{p}$ for $8 \times 8$ times (depicted as ``tiling'' in Figure~\ref{fig:generator}) and concatenate it with the other two tensors (depicted as ``channel-wise concatenation'' in Figure~\ref{fig:generator}). This makes the output of the convLSTM a $8 \times 8 \times (512+p)$ tensor at each time step; we set $p=96$ for the MUG dataset and $128$ for the NATOPS dataset.

The image decoder (the bottom two rows in Figure~\ref{fig:generator}) takes the concatenated output and produces the next frame by a series of deconvolutions. 
To avoid the checkerboard artifact in deconvolution~\cite{odena-distill16}, we use the upscale-convolution trick for all deconvolutional steps. 
The decoder architecture is \texttt{conv2d(256)} -- \texttt{BN} -- \texttt{leakyReLU} -- \texttt{upsample} -- \texttt{gating} -- \texttt{conv2d(128)} -- \texttt{BN} -- \texttt{leakyReLU} -- \texttt{upsample} -- \texttt{gating} -- \texttt{conv2d(64)} -- \texttt{BN} -- \texttt{leakyReLU} -- \texttt{upsample} -- \texttt{gating} -- \texttt{conv2d(64)} -- \texttt{BN} -- \texttt{leakyReLU} -- \texttt{conv2d(3)} -- \texttt{tanh}, where \texttt{conv2d($k$)} is a 2D convolutional layer with $k$ filters of $3 \times 3$ kernel with stride 1, and \texttt{upsample} is the $2 \times 2$ bilinear up-sampling. 

To provide a skip-connection from the image encoder to the decoder, we incorporate a \texttt{gating} operator that computes a weighted average of two tensors, whose weights are computed from the output of convLSTM at each time step. Specifically, we encode the convLSTM output with a small network of \texttt{upsample(2)} -- \texttt{conv2d(256)} -- \texttt{leakyReLU} -- \texttt{sigmoid} for \texttt{[Conv11]} , \texttt{upsample(4)} -- \texttt{conv2d(128)} -- \texttt{leakyReLU} -- \texttt{sigmoid} for \texttt{[Conv12]} and \texttt{upsample(8)} -- \texttt{conv2d(64)} -- \texttt{leakyReLU} -- \texttt{sigmoid} for \texttt{[Conv13]}, where \texttt{upsample($k$)} is a bilinear up-sampling of $k \times k$ kernel. The output of these small networks are used as weights for the gates. We then perform a weighted average of two tensors element-wise (depicted as ``element-wise gating'' in Figure~\ref{fig:generator}). Formally, denoting the output of the small network (e.g., sigmoid output of \texttt{[Conv11]}) by $s$, the result of the $2 \times 2$ upsampling (e.g., output of \texttt{[Upsample1]}) by $u$, and the tensor from the encoder via skip connection (e.g., output of \texttt{[Conv4]}) by $e$, the element-wise gating computes: $s \cdot u + (1-s) \cdot e$, where $\cdot$ is element-wise multiplication.

\begin{figure}[t]
    \centering
    \includegraphics[width=\linewidth]{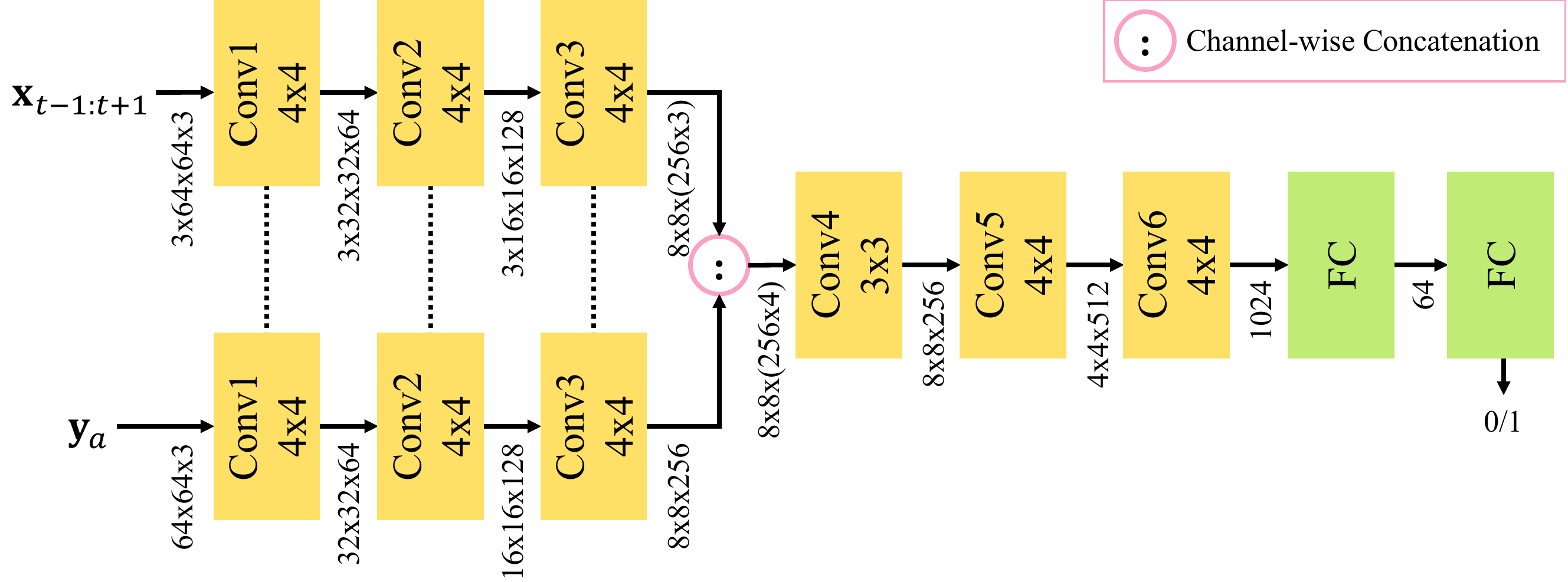}
    \caption{Appearance discriminator $\mathcal{D}_{a}$ network architecture.}
    \label{fig:appearance_discriminator}
\end{figure}

\subsection{Appearance Discriminator Network (Figure~\ref{fig:appearance_discriminator}) }

Our appearance discriminator takes four frame images as input: an appearance condition $\mathbf{y}_{a}$ (i.e., the first frame of a video) and three consecutive frames $\mathbf{x}_{t-1:t+1}$ from either a real or a generated video. It then outputs a scalar value indicating whether the quadruplet input is real or fake. 

We feed each image into a network of \texttt{conv2d(64,2)} -- \texttt{BN} -- \texttt{leakyReLU} -- \texttt{conv2d(128,2)} -- \texttt{BN} -- \texttt{leakyReLU} -- \texttt{conv2d(256,2)} -- \texttt{BN} -- \texttt{leakyReLU}, and concatenate the output from $\mathbf{y}_{a}$ and $\mathbf{x}_{t-1:t+1}$ channel-wise. We then take \texttt{deconv2d(256)} -- \texttt{BN} -- \texttt{leakyReLU} -- \texttt{conv2d(512,2)} -- \texttt{BN} -- \texttt{leakyReLU} -- \texttt{conv2d(1024,4)} -- \texttt{BN} -- \texttt{leakyReLU} -- \texttt{fc(64)} -- \texttt{BN} -- \texttt{leakyReLU} -- $\sigma(\texttt{fc(1)})$, where \texttt{conv2d($k$,$s$)} is a 2D convolutional layer with $k$ filters of $4 \times 4$ kernel with stride $s$, \texttt{deconv2d($k$)} is a 2D deconvolutional layer with $k$ filters of $3 \times 3$ kernel with stride 1 and \texttt{fc($k$)} is a fully-connected layer with $k$ units.

\begin{figure}[t]
    \centering
    \includegraphics[width=\linewidth]{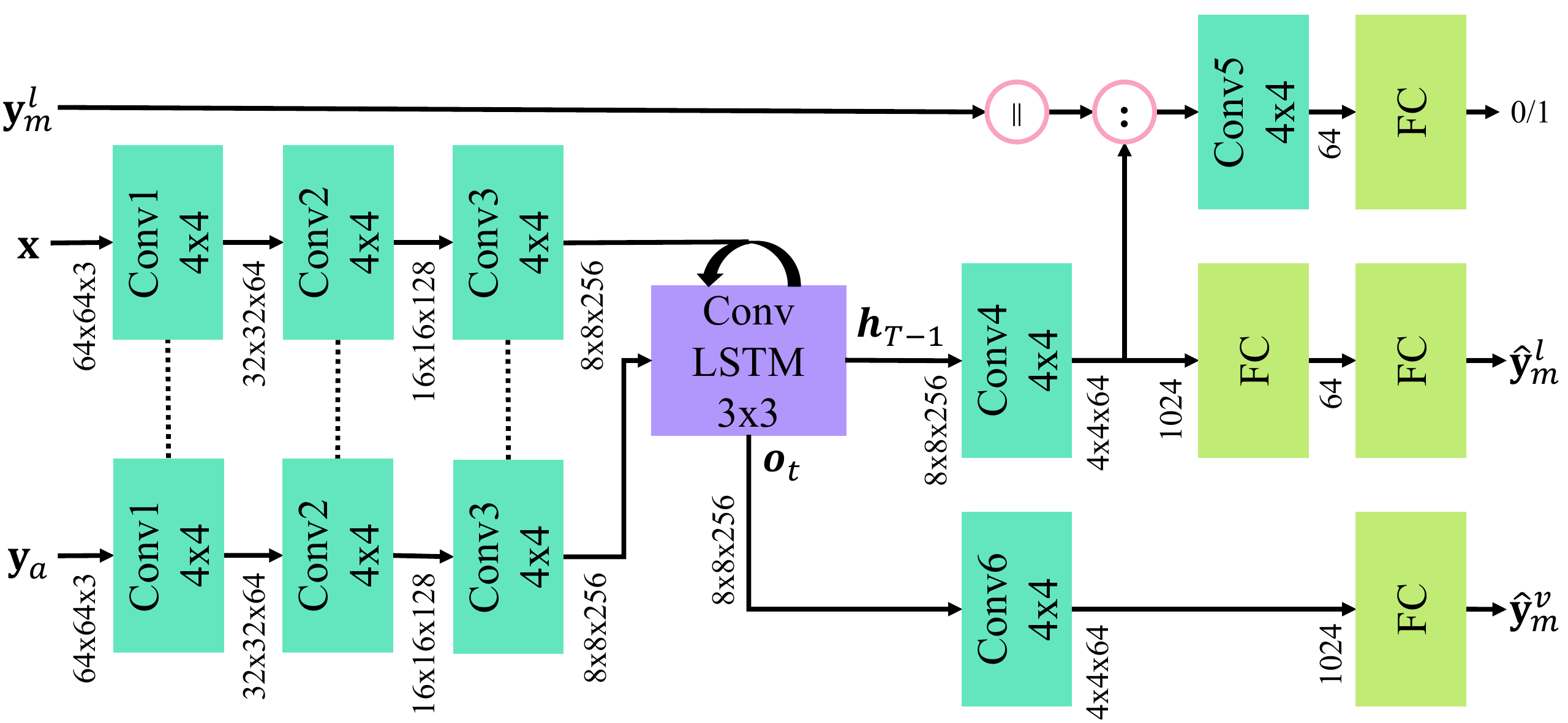}
    \caption{Motion discriminator $\mathcal{D}_{m}$ network architecture. Dashed lines indicate parameter sharing.}
    \label{fig:motion_discriminator}
\end{figure}

\subsection{Motion Discriminator Network (Figure~\ref{fig:motion_discriminator}) }

This network takes a video $\mathbf{x}$ with \textit{matched} appearance condition $\mathbf{y}_{a}$, and a motion class category $\mathbf{y}^{l}_{m}$ as input. It predicts three variables: a scalar indicating whether the video is real or fake, $\hat{y}_{m}^{l} \in \mathbb{R}^{c}$ representing motion categories, and $\hat{y}_{m}^{v} \in \mathbb{R}^{2k}$ representing the velocity of $k$ keypoints. 

We encode each frame of $\mathbf{x}$ and $\mathbf{y}_{a}$ with \texttt{conv2d(64)} -- \texttt{leakyReLU} -- \texttt{conv2d(128)} -- \texttt{BN} -- \texttt{leakyReLU} -- \texttt{conv2d(256)} -- \texttt{BN} -- \texttt{leakyReLU}, where \texttt{conv2d($k$)} is a 2D convolutional layer with $k$ filters of $4 \times 4$ kernel with stride of 2. We then use the encoded $\mathbf{y}_{a}$ to initialize the hidden states of the convLSTM, which has $256$ filters of $3 \times 3$ kernel with stride 1. At each time step $t$, we feed the encoded frame $\mathbf{x}_{t}$ to the convLSTM to produce output $\mathbf{o}_{t}$.

In each time step output $\mathbf{o}_{t}$, we feed it to \texttt{conv2d(64,2)} -- \texttt{BN} -- \texttt{leakyReLU} -- \texttt{flatten} -- \texttt{tanh(fc(\#points x 2))} to predict the velocity at each time step, where \texttt{conv2d($k,s$)} is a 2D convolutional layer with $k$ filters of $4 \times 4$ kernel with stride of $s$, \texttt{fc($k$)} is a fully-connected layer with $k$ units. Similarly, with the last hidden state $\mathbf{h}_{T-1}$,  we feed it to \texttt{conv2d(64,2)} -- \texttt{BN} -- \texttt{leakyReLU} -- \texttt{flatten} -- \texttt{fc(64)} -- \texttt{BN} -- \texttt{leakyReLU} -- \texttt{fc(\#class)} -- \texttt{BN} -- \texttt{leakyReLU} -- \texttt{softmax} to predict the motion class category. Also, for conditional prediction, we share the output of \texttt{conv2d(64,2)} -- \texttt{BN} -- \texttt{leakyReLU} step and concatenate the motion class after replicating them. After that, we feed it to \texttt{conv2d(64,4)} -- \texttt{BN} -- \texttt{leakyReLU} -- $\sigma(\texttt{fc(1)})$ to judge whether given video have matched motion or not.

\section{Experiment Details}

\begin{figure*}[t]
    \centering
    \includegraphics[width=0.4\linewidth]{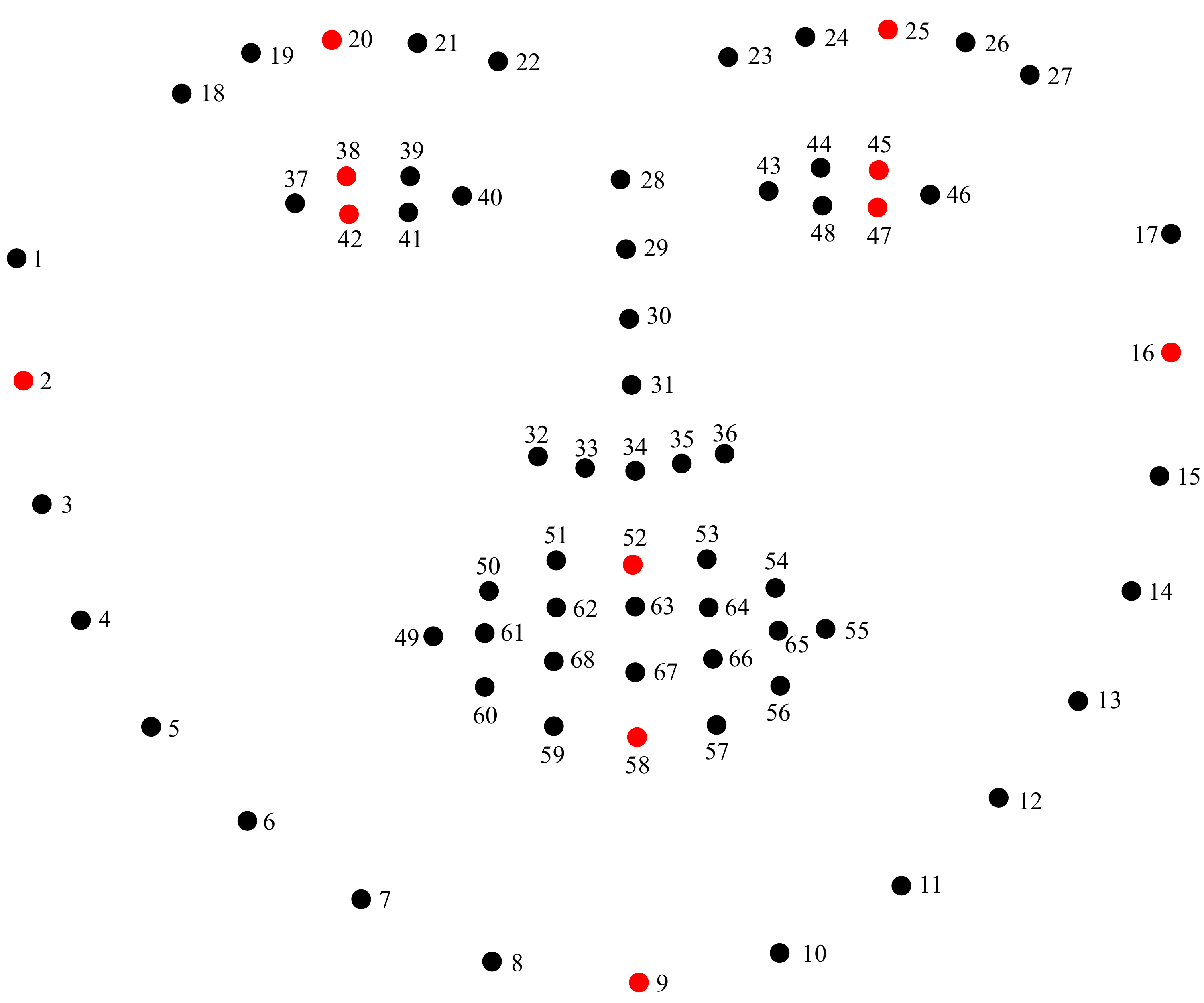}
\caption{The 68 facial-landmark template used by OpenFace~\cite{baltruvsaitis-wacv16}. We used 11 landmarks as facial keypoints (highlighted in red).}
    \label{fig:landmark_positions}
\end{figure*}

\begin{table*}[t]
\centering
\begin{tabular}{cl}
\hline
Base Emotions & EMFACS Prototypes \\\hline
Disgust & $9$ \\
& $9+16+25, 26$ \\
& $9+17$ \\
& $10^{*}$ \\
& $10^{*}+16+25,26$ \\
& $10+17$ \\\hline
Surprise & $1+2+5B+26,27$ \\
& $1+2+5B$ \\
& $1+2+26,27$ \\
& $5B+26,27$ \\\hline
Anger & $(4+5^{*}+7+10^{*}+22+23+25,26)^{**}$ \\
& $(4+5^{*}+7+10^{*}+23+25,26)^{**}$ \\
& $(4+5^{*}+7+23+25,26)^{**}$ \\
& $(4+5^{*}+7+17+23,24)^{**}$ \\
& $(4+5^{*}+7+23,24)^{**}$ \\\hline
Happiness & $6+12^{*}$ \\
 & $12C/D$ \\\hline
Sadness & $(1+4+11+15B +/- 54+64)+/- 25, 26$ \\
& $(1+4+15^{*} +/- 54+64)+/- 25, 26$ \\
& $(6+15^{*} +/- 54+64)+/- 25, 26$ \\
& $(1+4+15B +/- 54+64)+/- 25, 26$ \\
& $(1+4+15B+17 +/- 54+64)+/- 25, 26$ \\
& $(11+15B +/- 54+64)+/- 25, 26$ \\
& $11+17+/- 25, 26$ \\\hline
Fear & $1+2+4+5^{*}+20^{*}+ 25, 26,~or~27$ \\
& $1+2+4+5^{*}+ 25, 26,~or~27$ \\
& $1+2+4+5^{*}+L~or~R20^{*}+ 25, 26,~or~27$ \\
& $1+2+4+5^{*}$ \\
& $1+2+5Z, +/- 25, 26, 27$ \\
& $5^{*}+20^{*} +/- 25, 26, 27$ \\ \hline
\end{tabular}
\vspace{6pt}
\caption{The EMFACS (emotional facial action coding system) prototype table~\cite{fasel-sigmm04} that we used to select relevant action units (Section 4.1). * In this combination the AU may be at any level of intensity. ** Any of the prototypes can occur without any one of the following AUs: 4, 5, 7, or 10.}
\label{tab:emfacs_prototypes} 
\end{table*}

\subsection{Datasets}

\textbf{MUG facial expression}~\cite{aifanti-wiamis10}: The dataset contains 931 video clips performing six basic emotions~\cite{ekman-cogemo1992} (anger, disgust, fear, happy, sad, surprise). We preprocess it so that each video has 32 frames with 64 $\times$ 64 pixels (see below for details). For data augmentation we perform random horizontal flipping, and we use a random stride (1, 2, or 3) to sample frames around the ``peak'' frames. This results in 3,840 video clips, where we use 472 videos as test data.

For preprocessing, we use the OpenFace toolkit~\cite{baltruvsaitis-wacv16} to detect facial landmarks and action unit (AU) intensities (see Figure~\ref{fig:landmark_positions}). We consider only those AUs that are part of the EMFACS prototypes~\cite{friesen-unpub83}, listed in Table~\ref{tab:emfacs_prototypes}, under each video's ground truth emotion category. We identify one peak frame from each video that contains the maximum AU intensity (regardless of AU) and sample 32 frames around it (23 frames before and 8 after). Next, we use facial landmarks to center-align, rescale, and crop face regions to 64 $\times$ 64 pixels. We use 11 facial landmarks (2, 9, 16, 20, 25, 38, 42, 45, 47, 52, 58th) as the keypoints (shown in Figure~\ref{fig:landmark_positions}).

\textbf{NATOPS human action}~\cite{song-fg11}: The dataset consists of 9,600 video clips performing 24 action categories. We discard 765 clips that contain less than 32 frames, resulting in 8,835 clips; we use 1,810 clips as test data. We crop the video to 180 $\times$ 180 pixels with the chest at the center position and rescale it to 64 $\times$ 64 pixels. We use 9 joint locations (head, chest, naval, L/R-shoulders, L/R-elbows, L/R-wrists) available in the dataset as keypoints.

\subsection{The Loss Weights for Different Loss Terms}

Table~\ref{tab:hyperparameters} summarizes the loss weights for different loss terms in our model that we used for each dataset. 

\begin{table}[t]
\small
\centering
\begin{tabular}{|l||c|c|}
\hline
Loss term & MUG & NATOPS \\\hline
$\mathcal{L}_\mathit{gan}$ & 3.0 & 1.0 \\
$\mathcal{L}_\mathit{rank}$ & 100.0 & 100.0 \\
$\mathcal{L}_\mathit{CE}$ & 0.3 & 0.03  \\ 
$\mathcal{L}_\mathit{MSE}$ & 300.0 & 30.0 \\ 
$\|\mathbf{x}|_{\mathbf{y}} - \hat{\mathbf{x}}|_{\mathbf{y}}\|_{1}$ & 0.1 & 3.0 \\ 
$\sum\nolimits_{j} d_{j}(\mathbf{x}|_{\mathbf{y}}, \hat{\mathbf{x}}|_{\mathbf{y}})$ & 1.0 & 30.0 \\ 
\hline 
\end{tabular}
\caption{The loss weights for different loss terms to balance the effect of each term used in our experiments.}
\vspace{-0.5em}
\label{tab:hyperparameters} 
\end{table}

\begin{figure}[t]
    \centering
    \includegraphics[width=\linewidth]{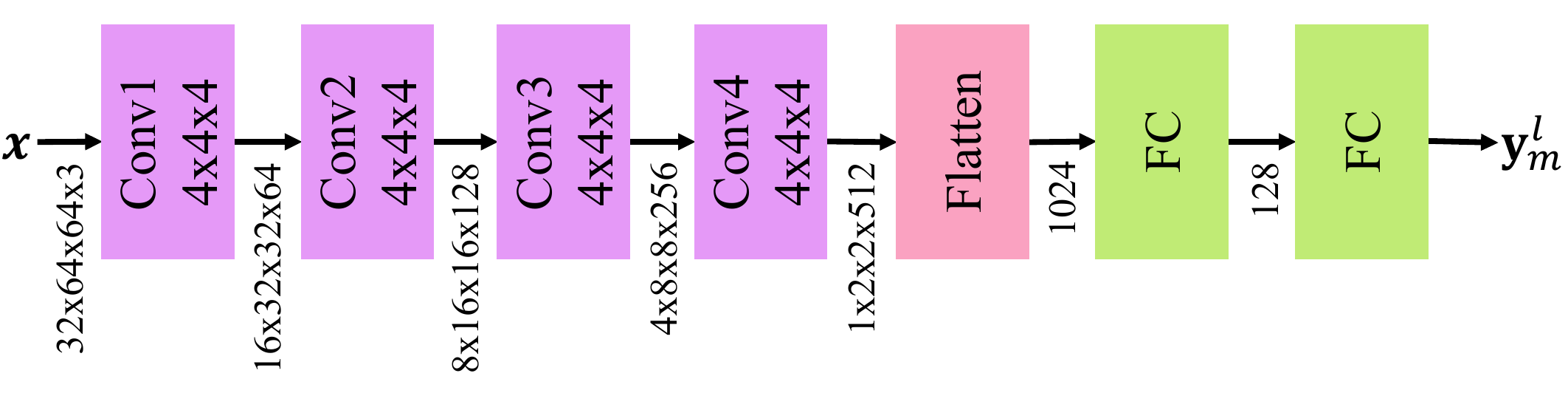}
    \caption{Image-based Motion predictor network architecture}
    \label{fig:image_based_motion_predictor}
\end{figure}

\subsection{3D CNN Motion Classifier (Figure~\ref{fig:image_based_motion_predictor})}

 We design a $c$-way motion classifier using a 3D CNN~\cite{tran-iccv15} that predicts the motion label $\mathbf{y}^{l}_{m}$ from a video. This network takes a video $\mathbf{x}$ as input and predicts a motion category label $\mathbf{y}^{l}_{m}$. To prevent the classifier from predicting the label simply by seeing the input frame(s), we discard the first four frames from the generated videos and use only the last 28 generated frames as input; we pad the first and last frame twice, respectively, and feed the 32 frames as input to the classifier.

We use a 3D CNN architecture of \texttt{conv3d(64)} -- \texttt{leakyReLU} -- \texttt{conv3d(128)} -- \texttt{BN} -- \texttt{leakyReLU} -- \texttt{conv3d(256)} -- \texttt{BN} -- \texttt{leakyReLU} -- \texttt{conv3d(512)}, where \texttt{BN} is batch normalization~\cite{ioffe-icml2015} and \texttt{conv3d($k$)} is a 3D convolutional layer~\cite{tran-iccv15} with $k$ filters of $4 \times 4 \times 4$ kernel. We use stride 2 for the first three \texttt{conv3d} layers and stride 4 for the last one. The output is an embedding $\phi(\mathbf{x})$ of size $2 \times 2 \times 512$. We flatten this network to the size of 2048 vectors and then feed it into \texttt{fc(128)} -- \texttt{BN} -- \texttt{leakyReLU} -- \texttt{dropout(0.5)} -- \texttt{fc($c$)} -- \texttt{softmax}. We set $c=6$ for the MUG facial expression dataset and $c=24$ for the NATOPS human action dataset.

\begin{figure}[t]
    \centering
    \includegraphics[width=\linewidth]{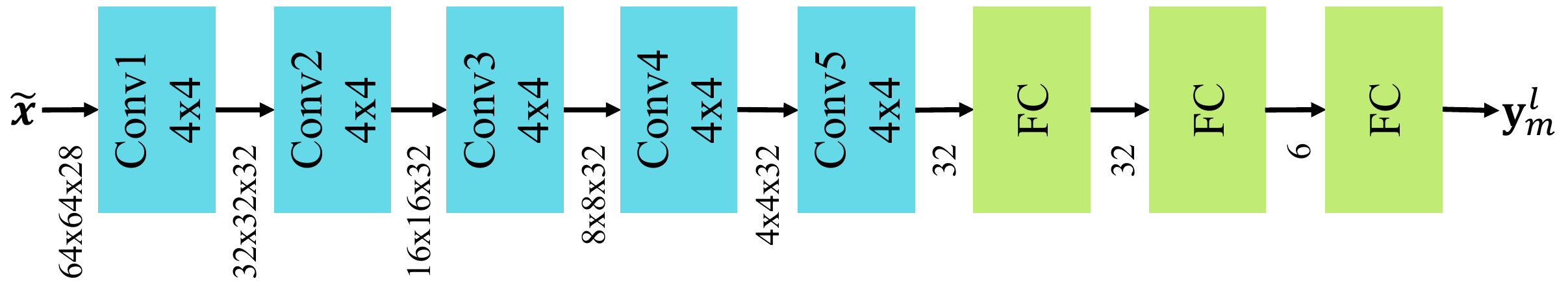}
    \caption{Keypoint-heatmap based motion predictor network architecture}
    \label{fig:keypoint_based_motion_predictor}
\end{figure}

\subsection{Keypoint-based Motion Predictor (Figure ~\ref{fig:keypoint_based_motion_predictor})}

This network takes a series of keypoint heatmaps $\tilde{\mathbf{x}}$, obtained from a video $\mathbf{x}$ as input and predicts the motion class category $\mathbf{y}^{l}_{m}$. Similar to the image-based motion predictor, we discard the first four frames from generated videos in order to avoid the video classifier learning to categorize them from the ground-truth frames in any case. 

To obtain the keypoint heatmaps, we first linearly upscale all real and generated videos to $128 \times 128$.  
We feed them to OpenFace~\cite{baltruvsaitis-wacv16} keypoint extractor, obtaining 68 keypoints for each frame. 
Then, we re-scale the keypoint coordinates so that they can fit into $64 \times 64$ frames (instead of $128 \times 128$). 
For each keypoint, We generate a Gaussian heatmap with the variance of 1/28.
Then, for each frame, we merge the 68 heatmaps ($64 \times 64$ pixels) into a single channel by taking the maximum value pixel-wisely.
The heatmaps for the last 28 frames are concatenated channel-wisely.

We use a 2D CNN architecture of \texttt{conv2d(32)} -- \texttt{BN} -- \texttt{leakyReLU} -- \texttt{conv2d(32)} -- \texttt{BN} -- \texttt{leakyReLU} -- \texttt{conv2d(32)} -- \texttt{BN} -- \texttt{leakyReLU} -- \texttt{conv2d(32)} -- \texttt{BN} -- \texttt{leakyReLU} -- \texttt{conv(32)} -- \texttt{BN} -- \texttt{leakyReLU}, where \texttt{BN} is batch normalization~\cite{ioffe-icml2015} and \texttt{conv2d($k$)} is a 2D convolutional layer~\cite{tran-iccv15} with $k$ filters of $4 \times 4$ kernel. We use stride 2 for the first four \texttt{conv2d} layers and stride 4 for the last one. The output is an embedding $\phi(\tilde{\mathbf{x}})$ of size $32$. We feed it into \texttt{fc(32)} -- \texttt{BN} -- \texttt{leakyReLU} -- \texttt{fc(6)}-- \texttt{BN}  -- \texttt{leakyReLU} -- \texttt{dropout(0.5)} -- \texttt{fc(6)} -- \texttt{softmax}.

\clearpage
\section*{Acknowledgements}
We thank Kang In Kim for helpful comments about building a human evaluation page. We also appreciate Youngjin Kim, Youngjae Yu, Juyoung Kim, Insu Jeon and Jongwook Choi for helpful discussions related to the design of our model. 
This work is partially supported by Korea-U.K. FP Programme through the National Research Foundation of Korea (NRF-2017K1A3A1A16067245).

{
\bibliographystyle{icml2018}
\bibliography{icml18_amcgan_arxiv}
}

\end{document}